\documentclass{article}
\usepackage{spconf,amsmath,graphicx}
\usepackage{amsfonts}
\usepackage{comment}
\usepackage{url}
\DeclareMathOperator*{\argmin}{argmin}

\title{A SMOKE REMOVAL METHOD for LAPAROSCOPIC IMAGES}
 \name{Congcong Wang$^{1,}$\sthanks{This work is funded by the Research Council of Norway through project no. 247689 IQ-MED: Image Quality enhancement in MEDical diagnosis, monitoring and treatment.} \qquad Faouzi Alaya Cheikh$^{1}$ \qquad Mounir Kaaniche$^{2}$\qquad Ole Jacob Elle$^{3,4}$}
\address{\normalsize$^{1}$ Norwegian Colour and Visual Computing Lab, Norwegian University of Science and Technology, Norway. \\
	\normalsize $^{2}$ L2TI-Institut Galil\'{e}e, Universit\'{e} Paris 13, Sorbonne Paris Cit\'{e}, Villetaneuse, France \\
	\normalsize $^{3}$The Intervention Centre, Oslo University Hospital, Oslo, Norway. \\
	\normalsize $^{4}$The Department of Informatics, University of Oslo, Oslo, Norway.
}
\begin{document}
%
\maketitle
\begin{abstract}
In laparoscopic surgery, image quality can be severely degraded by surgical smoke, which not only introduces error for the image processing (used in image guided surgery), but also reduces the visibility of the surgeons. In this paper, we propose to enhance the laparoscopic images by decomposing them into unwanted smoke part and enhanced part using a variational approach. 
The proposed method relies on the observation that smoke has low contrast and low inter-channel differences. A cost function is defined based on this prior knowledge and is solved using an augmented Lagrangian method. The obtained unwanted smoke component is then subtracted from the original degraded image, resulting in the enhanced image. The obtained quantitative scores in terms of FADE, JNBM and RE metrics show that our proposed method performs rather well. Furthermore, the qualitative visual inspection of the  results show that it removes smoke effectively from the laparoscopic images. 
\end{abstract}
\section{Introduction}
\label{sec:intro}
As one of the most important intra-operative data modality, laparoscopic images' quality 
is of vital importance for navigation systems and for the operating surgeons~\cite{stoyanov2012surgical}.
The artifacts during laparoscopic images specificities include smoke, blood, dynamic illumination conditions, specular reflections, \textit{etc}~\cite{sdiri2016adaptive}. Smoke significantly reduces the contrast and radiance information for large areas of the scene. Computer vision algorithms' performance and surgeons' visibility would inevitably suffer from this degradation. Therefore, smoke removal in laparoscopic images becomes necessary to improve the image guided surgery conditions and to provide a better operation field visualization. 
\par 
To the best of our knowledge, there is only a few recent works related to laparoscopic desmoking~\cite{kotwal2016joint, baid2017joint, tchakaa2017chromaticity, luo2017vision}. In these papers, the image desmoking problem is considered as a dehazing problem which has been studied for many years in the literature~\cite{tan2008visibility, he2011single}. In such problem, the atmospheric scattering model presented by Eq. (\ref{scattering}) describes the formation of a hazy image and is widely used in computer vision~\cite{narasimhan2002vision}.
\begin{small}
\begin{equation}
\label{scattering}
\mathbf{I}(x)=\mathbf{J}(x)t(x)+\mathbf{A}(1-t(x)),
\end{equation}
\end{small}
where $\mathbf{I}$ is the observed intensity, $\mathbf{J}$ is the scene radiance representing the haze-free image, $t$ is the medium transmission map considered to be inversely related to the scene's depth, and $\mathbf{A}$ is the airlight which is usually a constant as it is the global atmospheric light which is location independent.
\par 
However, while haze related to scene depth, smoke concentration is a local phenomenon which does not depend on the scene depth, but rather depends on the position of the tip of the thermal cutting instrument. Moreover, in laparoscopic images, the light source is provided from the instrument which is not evenly distributed, and the organ surface is not a Lambertian surface. These properties violate the assumptions underlying Eq. (\ref{scattering}), which makes it inappropriate for laparoscopic images.
\par 
In this paper, we propose a novel laparoscopic images smoke removal method. More precisely, instead of resorting to the classical physical atmospheric model, we propose another model where the degraded image is separated to two parts: the weighted smoke part and the desmoked one that should be recovered. To estimate the smoke, our approach relies on two easily verifiable assumptions: smoke has a low contrast and low inter-channel  differences.
\par 
The remainder of this paper is organized as follows. In Sec.~\ref{sec:rel_works}, a review of image dehazing as well as laparoscopic image desmoking methods are given. Sec.~\ref{sec:method} describes our proposed approach by defining the energy function and the optimization procedure. Finally, in Sec.~\ref{sec:result}, experimental results are presented and some conclusions are drawn in Sec.~\ref{sec:conclustion}.
\section{RELATED WORKS}
\label{sec:rel_works}
Recently, some works were proposed for desmoking in laparoscopic
images~\cite{kotwal2016joint, baid2017joint, tchakaa2017chromaticity, luo2017vision}. In~\cite{kotwal2016joint}, the authors formulated a joint desmoking and denoising problem as a Bayesian inference problem based on probabilistic graphical model. This work is then extended in~\cite{baid2017joint} for desmoking, denoising and specularity removal. 
In~\cite{tchakaa2017chromaticity}, an adapted dark-channel prior combined with histogram equalization method is presented. In~\cite{luo2017vision}, a visibility-driven fusion defogging framework is proposed.

While there is few works related to  laparoscopic images smoke removal, a similar problem referred to as image dehazing has also been studied in the literature. Many of the image dehazing works use the atmospheric scattering model and rely on the estimation of the transmission map $t$ or the depth map of the images and the airlight alternatively~\cite{zhu2015fast, he2011single, tarel2009fast}. He \textit{et al.} propose the dark channel approach based on a statistical observation from outdoor haze-free images: for most of the haze-free natural images, pixel values are very low for at least one channel~\cite{he2011single}. The transmission map $t$ computed by this prior together with an estimated $\mathbf{A}$ calculated from the detected most haze-opaque region of the image are applied to invert Eq. (\ref{scattering}), resulting in a haze free image. This is a well-known efficient approach and lots of recent methods are proposed based on it~\cite{tchakaa2017chromaticity, xu2012fast}. 
\par 
Some works have been also developed without estimating  transmission or depth maps. Tan \textit{et al.}~\cite{tan2008visibility} tried to enhance the haze image directly by maximizing the local contrast under an airlight smooth constraint.  
In~\cite{ancuti2013single}, a multi-scale fusion dehazing method is proposed by deriving a white balance and contrast enhanced inputs. 
\par 
In~\cite{galdran2014variational}, a variational contrast enhancement framework for image dehazing with a modified gray-world assumption is proposed. Later in~\cite{galdran2015enhanced}, an improved version is presented, where a saturation term is added to the variational cost function aiming to maximize the contrast and saturation together. In~\cite{galdran2017fusion}, Galdran \textit{et al.} further improved their work by enhancing faraway regions where normally have more fog and preserving nearby low-fog regions. Those methods do not rely on a physical atmospheric model, but try to maximize contrast and saturation. However, the modified gray-world assumption and the assumption that pixels' intensity is related to depth are violated in laparoscopic images. 
\par 

\par 
\section{Proposed smoke removal approach}
\label{sec:method}
Variational techniques have attracted a considerable attention over  the last years in signal/image processing literature. They have been found to be among the most powerful techniques in different fields such as  enhancement, restoration, super resolution and disparity/motion estimation from a sequence of images. For this reason, we propose here a variational approach to remove smoke in laparoscopic images. More precisely, an energy function is first defined and then minimized (i.e optimized) via an augmented Lagrangian method, as we shall address next.
\par 
\subsection{Energy function} 
Due to the aforementioned limitations of the atmospheric scattering model in laparoscopic images, we propose to consider another model where the degraded laparoscopic image is assumed to follow this decomposition strategy:
\begin{equation}
\label{model}
\mathbf{I}^{c}=\mathbf{J}^{c}+\alpha^{c} \cdot \mathbf{F}^{c},
\end{equation}
where $c\in\{R,G,B\} $ indicates the RGB channels, $\mathbf{I}$ is the obtained degraded image by laparoscopic camera, $\mathbf{J}$ contains the color image information, $\mathbf{F}$ is the unwanted smoke component and $\alpha^{c}$ is a scalar weight for every channel. Thus, the smoked images $\mathbf{I}$ is separated into two parts: the smoke part $\mathbf{F}$ and the enhanced one $\mathbf{J}$.
\par 
Based on the observations that smoke part's variation is smooth and the RGB channel differences are low as a result of the whitish property of smoke, we propose to estimate the smoke part by minimizing the following energy function:
\begin{equation}
\label{cost}
E=\frac{\lambda }{2}\left \| \mathbf{F}-\mathbf{I} \right \|^{2}+\left \| \mathbf{F}_{TV} \right \|_{2},
\end{equation}
where $\mathbf{I}$ is the degraded color image in the RGB color space, $\mathbf{F}$ is the smoke part to be estimated, $\lambda$ is a scalar to adjust weights between the two terms of the equation, and $\left \| \mathbf{F} _{TV}\right \|_{2}$ is an isotropic total variation (TV)-norm which is given by: 
\begin{equation}
\label{tv}
\left \| \mathbf{F} _{TV}\right \|_{2}=\sum_{i} \sqrt{\beta _{x}^2[\mathbf{D}_{x}\mathbf{F}]_{i}^{2}+\beta _{y}^2[\mathbf{D}_{y}\mathbf{F}]_{i}^{2}+\beta _{c}^2[\mathbf{D}_{c}\mathbf{F}]_{i}^{2}},
\end{equation}
where $\beta_{x}$, $\beta_{y}$, $\beta_{c}$ are three scalar parameters to balance the weights between the gradient of the color image and the inter-channel differences. $\mathbf{D}_{x}$, $\mathbf{D}_{y}$, $\mathbf{D}_{c}$ are the forward differential operators along the three dimensions. Thus, we have  $\mathbf{D}_{x}\mathbf{F}=\mathbf{F}(x+1,y,c)-\mathbf{F}(x,y,c)$, $\mathbf{D}_{y}\mathbf{F}=\mathbf{F}(x,y+1,c)-\mathbf{F}(x,y,c)$, and  $\mathbf{D}_{c}\mathbf{F}=\mathbf{F}(x,y,c+1)-\mathbf{F}(x,y,c)$. Note that $(x,y,c)$ represents the pixel coordinates of the color image with horizontal and vertical directions $(x,y)$ and channel direction $c$. Using matrix-vector notation, $[\mathbf{D}_{d}\mathbf{F}]_i$, with $d \in \{x,y,c\}$, denotes the $i$-th component of the one dimensional vector obtained from $\mathbf{D}_{d}\mathbf{F}$.  
\par 
The first term in Eq. (\ref{cost}) aims to keep the similarity between the estimated smoke part and the input degraded image. The second total variation norm part represents the properties of the smoke part: low contrast and low inter-channel differences.
\par 
After the estimation of global smoke $\textbf{F}$, the smoke free image $\textbf{J}$ is then calculated as:
\begin{equation}
\label{final}
\mathbf{J}^{c}=\mathbf{I}^{c}-\alpha^{c} \cdot \mathbf{F}^{c},
\end{equation}
where $\alpha^{c}$ is defined as the mean values of the estimated smoke image over the RGB channels.
\par 
\par 
\subsection{Optimization method}
The energy function minimization problem can be solved by employing the augmented Lagrangian method~\cite{chan2011augmented}, which will be described in the following. The function, given by Eq. (\ref{cost}), is split by introducing an intermediate new variable $\textbf{u}$:
\begin{equation}
\label{spliting}
\begin{split}
&\displaystyle{\min_{\textbf{F}}  \quad \frac{\lambda }{2}\left \| \mathbf{F}-\mathbf{I} \right \|^{2}+\left \| \mathbf{u} \right \|_{2}}, \\
&\textrm{\textit{s. t.}} \quad \mathbf{F}_{TV}-\mathbf{u}=0.
\end{split}
\end{equation}
The augmented Lagrangian for Eq. (\ref{spliting}) is:
\begin{small}
	\begin{equation}
	\label{split}
	L_{\rho}(\mathbf{F},\mathbf{u},\mathbf{y})=\frac{\lambda }{2}\left \| \mathbf{F}-\mathbf{I} \right \|^{2}+\left \| \mathbf{u} \right \|_{2}+\mathbf{y}^{T}(\mathbf{F}_{TV}-\mathbf{u})+\frac{\rho}{2}\left \|\mathbf{F} _{TV}- \mathbf{u}\right \|^{2},
	\end{equation}
\end{small}
where $\rho$ is a non-negative constant parameter called penalty parameter and $\mathbf{y}=[\mathbf{y}_{x}^{\top},\mathbf{y}_{y}^{\top},\mathbf{y}_{c}^{\top}]^{\top}$ is the Lagrange multipliers vector and $\mathbf{u}=[\mathbf{u}_{x}^{\top},\mathbf{u}_{y}^{\top},\mathbf{u}_{c}^{\top}]^{\top}$. Then, the alternating direction method (ADM)~\cite{boyd2011distributed} is used to solve the following minimization sub-problems iteratively: 
\begin{footnotesize}
	\begin{equation}
	\label{subproblem}
	\begin{split}
	\textbf{F}^{k+1}&:=\argmin_{\textbf{F}} L_{\rho}(\textbf{F},\textbf{u}^{k},\textbf{y}^{k}),\\
	&=\argmin_{\textbf{F}} \frac{\lambda }{2}\left \| \mathbf{F}-\mathbf{I} \right \|^{2}+(\mathbf{y}^{k})^{\top}(\mathbf{F}_{TV}-\mathbf{u}^{k})+\frac{\rho}{2}\left \|\mathbf{F} _{TV}- \mathbf{u}^{k}\right \|^{2},  \\
	\textbf{u}^{k+1}&:=\argmin_{\textbf{u}} L_{\rho}(\textbf{F}^{k+1},\textbf{u},\textbf{y}^{k}),\\
	&=\argmin_{\textbf{u}} \left \| \mathbf{u} \right \|_{2}+ (\mathbf{y}^{k})^{\top}(\mathbf{F}_{TV}^{k+1}-\mathbf{u})+\frac{\rho}{2}\left \|\mathbf{F} _{TV}^{k+1}- \mathbf{u}\right \|^{2},  \\
	\textbf{y}^{k+1}&:= \textbf{y}^{k}+\rho(\textbf{F}_{TV}^{k+1}-\textbf{u}^{k+1}).
	\end{split}
	\end{equation}
\end{footnotesize}
By introducing the operator $\textbf{D}=[\beta _{x}\textbf{D}_{x}^{\top},\beta _{y}\textbf{D}_{y}^{\top},\beta _{c}\textbf{D}_{c}^{\top}]^{\top}$, the $\textbf{F}$-minimization subproblem leads to the following solution:
\begin{equation}
\label{fproblem}
\mathbf{F}=\mathcal{F}^{-1}[\frac{\mathcal{F}[\lambda \mathbf{I}+\rho\mathbf{D}^{\top}\mathbf{u}-\mathbf{D}^{\top}\mathbf{y}]}{\lambda+\rho(\left | \beta_{x}\mathcal{F}[\textbf{D}_{x}]\right |^{2}+\left |\beta_{y}\mathcal{F}[\textbf{D}_{y}]\right |^{2}+\left|\beta_{c}\mathcal{F}[\textbf{D}_{c}]\right |^{2} )}],
\end{equation}
where $\mathcal{F}$ is the Fourier transform operator. Then, the
$\textbf{u}$ minimization subproblem results in:
\begin{equation}
\label{uproblem}
\mathbf{u}_{x}=\mbox{max}\left \{  \mathbf{v}-\frac{1}{\rho},0 \right \} \cdot \frac{\mathbf{v}_{x}}{\mathbf{v}},
\end{equation}
where $\mathbf{v}_{x}=\beta_{x} \textbf{D}_{x}\textbf{F}+(\frac{1}{\rho})\textbf{y}_{x}$. Similar definition is applied to $\textbf{v}_{y}$, $\textbf{v}_{c}$, and $\textbf{v}=\mbox{max} \{ \sqrt{\left|\textbf{v}_{x}\right|^{2}+\left|\textbf{v}_{y}\right|^{2} +\left|\textbf{v}_{c}\right|^{2}}, \epsilon  \}$ with $\epsilon $ a small constant. In a similar way, $\textbf{u}_{y}$ and $\textbf{u}_{c}$ are determined to obtain the vector $\mathbf{u}$. More details about these solutions can be found in~\cite{chan2011numerical}.


\section{EXPERIMENTAL RESULTS}
\label{sec:result}
In vivo procedure datasets~\cite{giannarou2013probabilistic, ye2017self}, taken from Hamlyn Centre Laparoscopic / Endoscopic Video Dataset Page\footnote{http://hamlyn.doc.ic.ac.uk/vision/}, are used for validation. 
\textit{Dataset1} has 96 smoked images and \textit{Dataset2} contains 4031 images. In order to show the benefits of the proposed method, we will compare it to the following recent ones~\cite{he2011single,galdran2015enhanced, galdran2017fusion}. The first one is the atmospheric model based image dehazing method with dark channel prior~\cite{he2011single}. This method will be designated by DCP. It is important to note here that similar approach has been investigated in~\cite{tchakaa2017chromaticity} to remove smoke by adding thresholding or refining steps. However, the later approach has not been considered in our evaluation because of its sensitivity to different parameters which should be empirically selected for input smoked images of the large experimental dataset.  
The second one, which will be denoted by E-VAR, corresponds to an enhanced variational approach developed in~\cite{galdran2015enhanced}. Finally, the third one, designated by F-VAR, is a fusion-based variational technique~\cite{galdran2017fusion}. 
The parameters setting used in this experiment are: $\lambda=1$ for Eq.~(\ref{cost}), $\beta_{x}=\beta_{y}=\beta_{c}=1$ for Eq.~(\ref{tv}) and $\rho=5$. 
\begin{table}[t]
	\resizebox{.48\textwidth}{!}{
		\begin{centering}  %
			\begin{tabular}{ccccc}  
				\hline
				\hline
				&FADE~\cite{choi2015referenceless}&JNBM~\cite{ferzli2009no}&RE~\cite{hautiere2011blind}\\ \hline 
				Input images 						&$0.40\pm 0.03 $ &$1.42\pm 0.12 $&$0$& \\
				DCP~\cite{he2011single}     		&$0.27\pm 0.01$  &$1.57\pm 0.14 $&$0.38\pm 0.06$&\\
				F-VAR~\cite{galdran2017fusion}   	&$0.43\pm 0.02$  & $1.62\pm 0.12$&$0.12\pm 0.02$&\\
				E-VAR~\cite{galdran2015enhanced} 	&$0.35\pm 0.02$  &$1.50\pm 0.11$&$0.24\pm 0.05$&\\
				Proposed 							&$\mathbf{0.23\pm0.02}$&$\mathbf{1.77 \pm 0.11}$&$\mathbf{0.39 \pm 0.07}$&\\ \hline \hline
			\end{tabular}
		\end{centering}
	}
	\caption{Quantitative evaluation results for \textit{Dataset1}.}
	\label{result1}
\end{table}
\begin{table}[t]
	\resizebox{.48\textwidth}{!}{
		\begin{centering}  %
			\begin{tabular}{ccccc}  
				\hline \hline
				&FADE~\cite{choi2015referenceless}&JNBM~\cite{ferzli2009no}&RE~\cite{hautiere2011blind}\\ \hline 
				Input images 					&$0.67 \pm0.16$	 &$1.03\pm 0.11$ & $0$& \\
				DCP~\cite{he2011single}     	&$0.33 \pm 0.05$ &$1.06\pm 0.11$  &$0.88\pm 0.42$&\\
				F-VAR~\cite{galdran2017fusion}  &$0.50 \pm 0.09$ & $1.09\pm 0.11$&$0.41\pm 0.20$&\\
				E-VAR~\cite{galdran2015enhanced}&$0.36 \pm 0.05$ &$1.05\pm 0.10$ &$0.73\pm 0.39$&\\
				Proposed						&$\mathbf{0.30\pm 0.05}$&$\mathbf{1.16 \pm 0.10}$&$\mathbf{1.19\pm 0.62}$&\\ \hline \hline
			\end{tabular}
		\end{centering}
	}
	\caption{Quantitative evaluation results for \textit{Dataset2}.}
	\label{result2}
\end{table}
\begin{figure*}[htb]
	\begin{minipage}[b]{0.19\linewidth}
		\centering
		\centerline{\includegraphics[width=3.3cm]{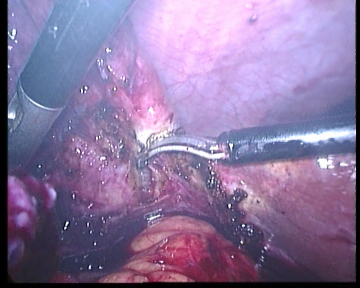}}
		\centerline{ }
	\end{minipage}
	\begin{minipage}[b]{.19\linewidth}
		\centering
		\centerline{\includegraphics[width=3.3cm]{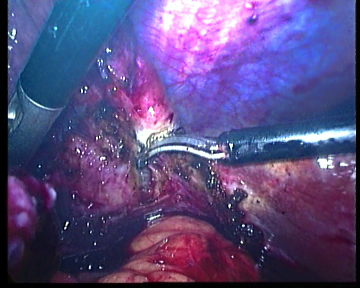}}
		\centerline{}
	\end{minipage}
	\begin{minipage}[b]{0.19\linewidth}
		\centering
		\centerline{\includegraphics[width=3.3cm]{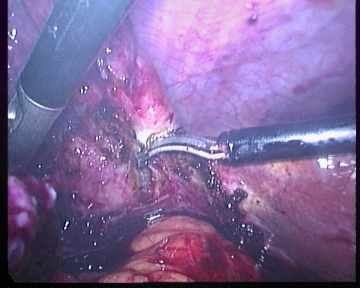}}
		\centerline{}
	\end{minipage}
	\begin{minipage}[b]{0.19\linewidth}
		\centering
		\centerline{\includegraphics[width=3.3cm]{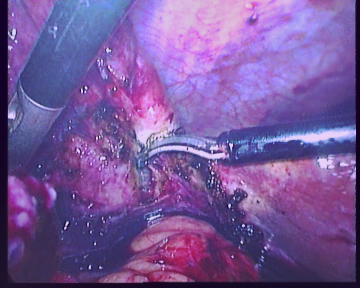}}
		\centerline{}
	\end{minipage}
	\begin{minipage}[b]{0.19\linewidth}
		\centering
		\centerline{\includegraphics[width=3.3cm]{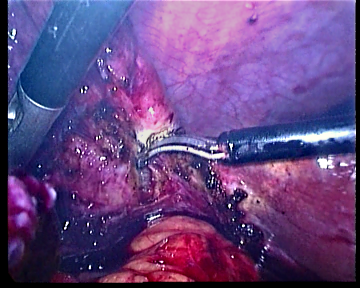}}
		\centerline{}
	\end{minipage}
	
	\begin{minipage}[b]{0.19\linewidth}
		\centering
		\centerline{\includegraphics[width=3.3cm]{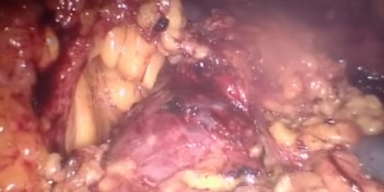}}
		\centerline{ }
	\end{minipage}
	\begin{minipage}[b]{.19\linewidth}
		\centering
		\centerline{\includegraphics[width=3.3cm]{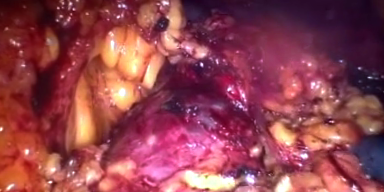}}
		\centerline{}
	\end{minipage}
	\begin{minipage}[b]{0.19\linewidth}
		\centering
		\centerline{\includegraphics[width=3.3cm]{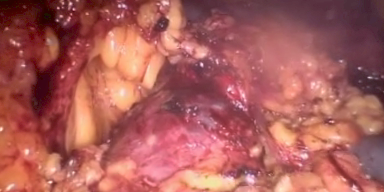}}
		\centerline{ }
	\end{minipage}
	\begin{minipage}[b]{0.19\linewidth}
		\centering
		\centerline{\includegraphics[width=3.3cm]{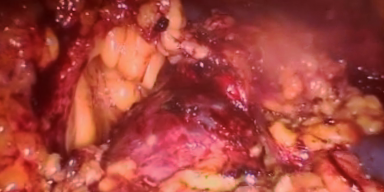}}
		\centerline{}
	\end{minipage}
	\begin{minipage}[b]{0.19\linewidth}
		\centering
		\centerline{\includegraphics[width=3.3cm]{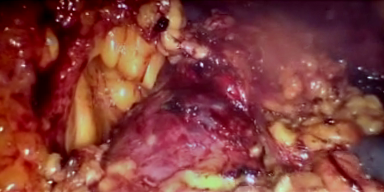}}
		\centerline{ }
	\end{minipage}
	
	\begin{minipage}[b]{0.19\linewidth}
		\centering
		\centerline{\includegraphics[width=3.3cm]{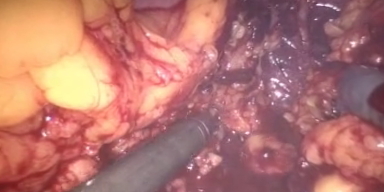}}
		\centerline{(a) }
	\end{minipage}
	\begin{minipage}[b]{.19\linewidth}
		\centering
		\centerline{\includegraphics[width=3.3cm]{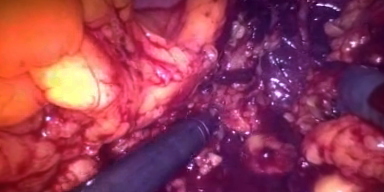}}
		\centerline{(b)}
	\end{minipage}
	\begin{minipage}[b]{0.19\linewidth}
		\centering
		\centerline{\includegraphics[width=3.3cm]{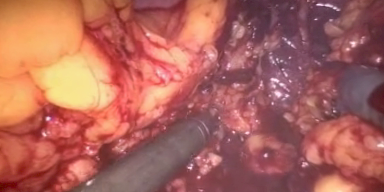}}
		\centerline{(c) }
	\end{minipage}
	\begin{minipage}[b]{0.19\linewidth}
		\centering
		\centerline{\includegraphics[width=3.3cm]{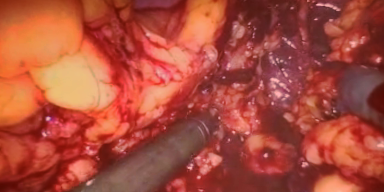}}
		\centerline{(d) }
	\end{minipage}
	\begin{minipage}[b]{0.19\linewidth}
		\centering
		\centerline{\includegraphics[width=3.3cm]{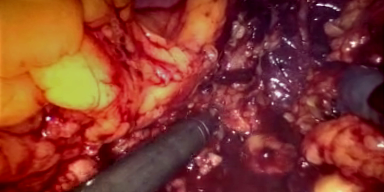}}
		\centerline{(e) }
	\end{minipage}
	\caption{Subjective results for \textit{Dataset1} and \textit{Dataset2}. (a) Input smoked laparoscopic image and the obtained desmoked ones using: (b) DCP~\cite{he2011single}, (c) F-VAR~\cite{galdran2017fusion}, (d) E-VAR~\cite{galdran2015enhanced}, and (e) proposed method.}
	\label{fig:res1}
\end{figure*}
\begin{figure*}[htb]
	\begin{minipage}[b]{0.33\linewidth}
		\centering
		\centerline{\includegraphics[width=1.1\linewidth]{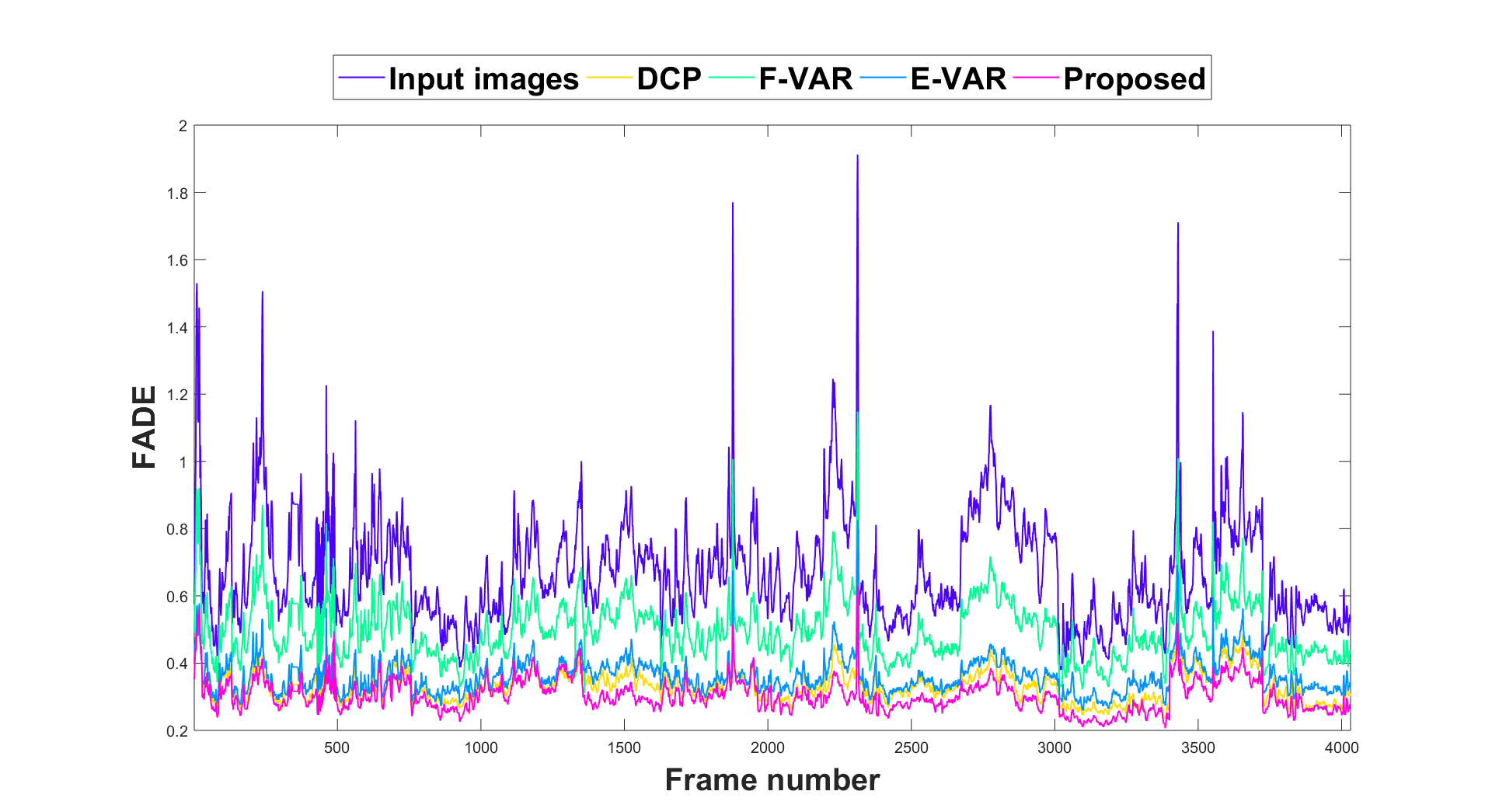}}
		\centerline{ (a) }
	\end{minipage}
	\begin{minipage}[b]{.33\linewidth}
		\centering
		\centerline{\includegraphics[width=1.1\linewidth]{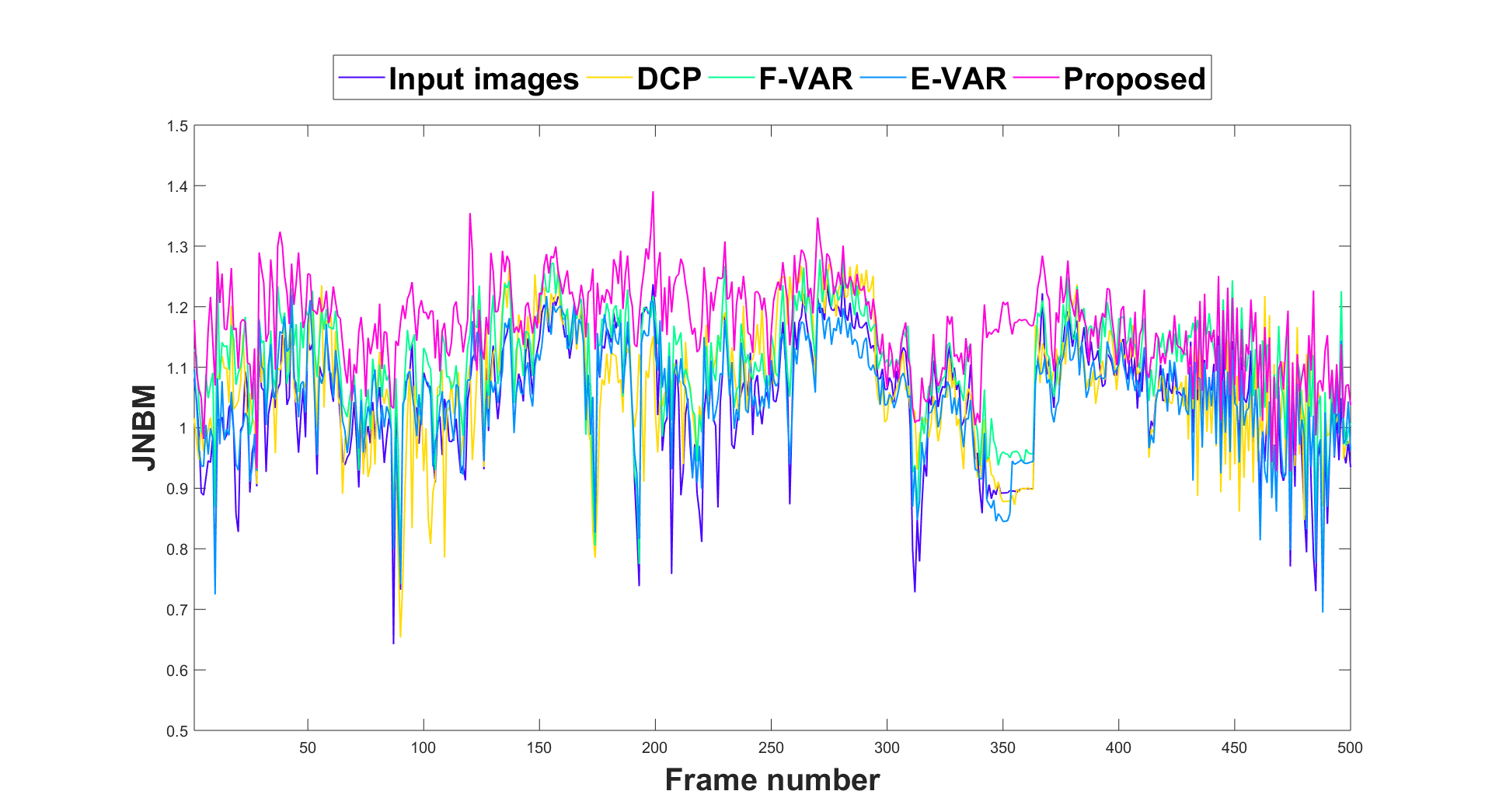}}
		\centerline{ (b) }
	\end{minipage}
	\begin{minipage}[b]{0.33\linewidth}
		\centering
		\centerline{\includegraphics[width=1.1\linewidth]{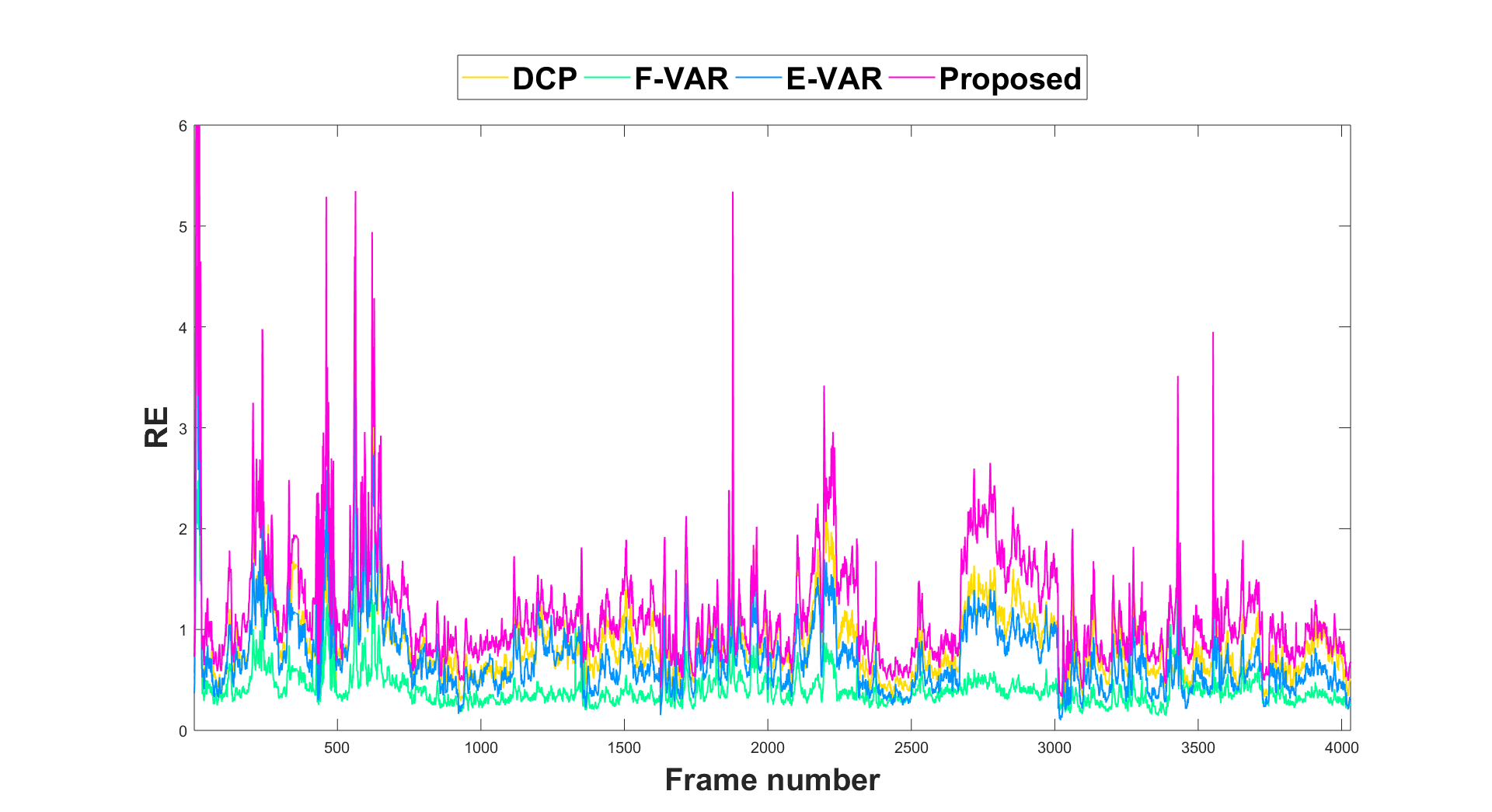}}
		\centerline{ (c) }
	\end{minipage}
	\caption{Plotted metrics for \textit{Dataset2}. (a) FADE~\cite{choi2015referenceless}, (b) JNBM~\cite{ferzli2009no}, (c) RE~\cite{hautiere2011blind}. Note that, for the JNBM, only 500 frames are plotted to provide better illustration.}
	\label{fig:res2}
\end{figure*}
\par 
\noindent
\textbf{Quantitative evaluation:}
Examples of three images are shown in Fig.~\ref{fig:res1}(a). 
As the ground-truth information for a smoked laparoscopic image is not available, we propose to employ two no-reference image quality metrics and another metric that compares the visibility of edges before and after smoke removal. For the purpose of evaluating the ability of smoke removal, a referenceless Fog Aware Density Evaluator (FADE) has been used to evaluate the perceptual fog density~\cite{choi2015referenceless, fog}. A lower FADE value means a lower perceptual fog density.
Besides, a just noticeable blur based no-reference objective image sharpness metric (JNBM)~\cite{ferzli2009no} is used to evaluate the perceptual sharpness. A higher value means higher perceptual sharpness or lower blurriness. 
Further more, we employ a metric, proposed by Hauti\`ere \textit{et al.}~\cite{hautiere2011blind}, which aims to assess the ability of restoring edges (RE) that are not visible in \textbf{I} but are in \textbf{J} (obtained after smoke removal). This metric will be designated by RE. A higher RE value means a better edge restoration.  
\par 
Tables~\ref{result1} and~\ref{result2} show mean and standard deviation of the scores of the different approaches for \textit{Dataset1} and \textit{Dataset2}. Fig.~\ref{fig:res2} illustrates the scores of the three metrics on \textit{Dataset2}. All the three metrics show better scores for our approach. In terms of  FADE metric, the DCP method removes smoke well. However, it scarifies the perceptual quality as shown in Fig.~\ref{fig:res1}~ (b) as a result of the constant airlight assumption. E-VAR removes more smoke than F-VAR. F-VAR's results indicate that there are still high smoke density in the images. Our proposed method's smoke density is the lowest. The proposed approach removes the smooth smoke component of the image resulting in a contrast enhanced image, which has the best scores for JNBM and RE. 
\par 
\noindent
\textbf{Qualitative evaluation:} In this part, we evaluate the different methods subjectively. 
Fig.~\ref{fig:res1} illustrates the results for three laparoscopic images from the two datasets. It can be observed, from Fig.~\ref{fig:res1}(b), that the DCP method can remove the smoke effectively but causes an unnatural color change in the desmoked images. Moreover, as shown in Fig.~\ref{fig:res1}(c), the smoke is not well removed by the F-VAR approach as smoke is independent of depth in laparoscopic images and this approach tries to preserve image information on nearby region under the assumption that the concentration of the haze is related with depth.
E-VAR~\cite{galdran2015enhanced} method relies mildly on the physical model, leads to the fine result shown in Fig.~\ref{fig:res1}(d).  
Finally, Fig.~\ref{fig:res1}(e) shows that our proposed method allows to remove smoke effectively, leading to an output image with enhanced contrast. 
\par 
Therefore, all the obtained results confirm the benefits of the proposed desmoking method for laparoscopic images.
\section{Conclusion}
\label{sec:conclustion}
Unlike most of the natural image dehazing methods which rely on a physical model, we propose in this paper a variational desmoking method for laparoscopic images. The aim is to remove the smoke from the scene, thus to improve the image guided surgery condition as well as the surgeons' visibility. Quantitative and qualitative evaluations are performed and show that the proposed approach reduces the smoke effectively while preserving the important perceptual information of the image. Further work should include a more robust prior about the smoke.

\begin{small}
\bibliographystyle{IEEEbib}
\bibliography{strings,refs}

\begin{thebibliography}{10}

\bibitem{stoyanov2012surgical}
D.~Stoyanov,
\newblock ``Surgical vision,''
\newblock {\em Annals of biomedical engineering}, vol. 40, no. 2, pp. 332--345,
  2012.

\bibitem{sdiri2016adaptive}
B.~Sdiri, A.~Beghdadi, F.~A. Cheikh, M.~Pedersen, and O.~J. Elle,
\newblock ``An adaptive contrast enhancement method for stereo endoscopic
  images combining binocular just noticeable difference model and depth
  information,''
\newblock {\em Electronic Imaging}, vol. 2016, no. 13, pp. 1--7, 2016.

\bibitem{kotwal2016joint}
A.~Kotwal, R.~Bhalodia, and S.~P. Awate,
\newblock ``Joint desmoking and denoising of laparoscopy images,''
\newblock in {\em IEEE international Symposium on Biomedical Imaging (ISBI)},
  2016, pp. 1050--1054.

\bibitem{baid2017joint}
A.~Baid, A.~Kotwal, R.~Bhalodia, S.~N. Merchant, and S.~P. Awate,
\newblock ``Joint desmoking, specularity removal, and denoising of laparoscopy
  images via graphical models and bayesian inference,''
\newblock in {\em IEEE international Symposium on Biomedical Imaging (ISBI)},
  2017, pp. 732--736.

\bibitem{tchakaa2017chromaticity}
K.~Tchakaa, V.~M. Pawara, and D.~Stoyanova,
\newblock ``Chromaticity based smoke removal in endoscopic images,''
\newblock in {\em Proc. of SPIE Vol}, 2017, vol. 10133, pp. 101331M--1.

\bibitem{luo2017vision}
X.~Luo, A.~McLeod, S.~Pautler, C.~Schlachta, and T.~Peters,
\newblock ``Vision-based surgical field defogging,''
\newblock {\em IEEE Transactions on Medical Imaging}, 2017.

\bibitem{tan2008visibility}
R.~T. Tan,
\newblock ``Visibility in bad weather from a single image,''
\newblock in {\em IEEE International Conference on Computer Vision and Pattern
  Recognition (CVPR)}, 2008, pp. 1--8.

\bibitem{he2011single}
K.~He, J.~Sun, and X.~Tang,
\newblock ``Single image haze removal using dark channel prior,''
\newblock {\em IEEE transactions on pattern analysis and machine intelligence},
  vol. 33, no. 12, pp. 2341--2353, 2011.

\bibitem{narasimhan2002vision}
S.~G. Narasimhan and S.~K. Nayar,
\newblock ``Vision and the atmosphere,''
\newblock {\em International Journal of Computer Vision}, vol. 48, no. 3, pp.
  233--254, 2002.

\bibitem{zhu2015fast}
Q.~Zhu, J.~Mai, and L.~Shao,
\newblock ``A fast single image haze removal algorithm using color attenuation
  prior,''
\newblock {\em IEEE Transactions on Image Processing}, vol. 24, no. 11, pp.
  3522--3533, 2015.

\bibitem{tarel2009fast}
J.~P. Tarel and N.~Hautiere,
\newblock ``Fast visibility restoration from a single color or gray level
  image,''
\newblock in {\em IEEE International Conference on Computer Vision (ICCV)},
  2009, pp. 2201--2208.

\bibitem{xu2012fast}
H.~Xu, J.~Guo, Q.~Liu, and L.~Ye,
\newblock ``Fast image dehazing using improved dark channel prior,''
\newblock in {\em IEEE International Conference on Information Science and
  Technology (ICIST)}, 2012, pp. 663--667.

\bibitem{ancuti2013single}
C.~O. Ancuti and C.~Ancuti,
\newblock ``Single image dehazing by multi-scale fusion,''
\newblock {\em IEEE Transactions on Image Processing}, vol. 22, no. 8, pp.
  3271--3282, 2013.

\bibitem{galdran2014variational}
A.~Galdran, J.~Vazquez-Corral, D.~Pardo, and M.~Bertalm{\'\i}o,
\newblock ``A variational framework for single image dehazing,''
\newblock in {\em European Conference on Computer Vision (ECCV) Workshops (3)},
  2014, pp. 259--270.

\bibitem{galdran2015enhanced}
A.~Galdran, J.~Vazquez-Corral, D.~Pardo, and M.~Bertalm{\'\i}o,
\newblock ``Enhanced variational image dehazing,''
\newblock {\em SIAM Journal on Imaging Sciences}, vol. 8, no. 3, pp.
  1519--1546, 2015.

\bibitem{galdran2017fusion}
A.~Galdran, J.~Vazquez-Corral, D.~Pardo, and M.~Bertalm{\'\i}o,
\newblock ``Fusion-based variational image dehazing,''
\newblock {\em IEEE Signal Processing Letters}, vol. 24, no. 2, pp. 151--155,
  2017.

\bibitem{chan2011augmented}
S.~H. Chan, R.~Khoshabeh, K.~B. Gibson, P.~E. Gill, and T.~Q. Nguyen,
\newblock ``An augmented lagrangian method for total variation video
  restoration,''
\newblock {\em IEEE Transactions on Image Processing}, vol. 20, no. 11, pp.
  3097--3111, 2011.

\bibitem{boyd2011distributed}
S.~Boyd, N.~Parikh, E.~Chu, B.~Peleato, and J.~Eckstein,
\newblock ``Distributed optimization and statistical learning via the
  alternating direction method of multipliers,''
\newblock {\em Foundations and Trends{\textregistered} in Machine Learning},
  vol. 3, no. 1, pp. 1--122, 2011.

\bibitem{chan2011numerical}
H.~Chan,
\newblock {\em Numerical optimization for image and video restoration},
\newblock University of California, San Diego, 2011.

\bibitem{giannarou2013probabilistic}
S.~Giannarou, M.~Visentini-Scarzanella, and G.~Z. Yang,
\newblock ``Probabilistic tracking of affine-invariant anisotropic regions,''
\newblock {\em IEEE transactions on pattern analysis and machine intelligence},
  vol. 35, no. 1, pp. 130--143, 2013.

\bibitem{ye2017self}
M.~Ye, E.~Johns, A.~Handa, L.~Zhang, P.~Pratt, and G.~Z. Yang,
\newblock ``Self-supervised siamese learning on stereo image pairs for depth
  estimation in robotic surgery,''
\newblock {\em arXiv preprint arXiv:1705.08260}, 2017.

\bibitem{choi2015referenceless}
L.~K. Choi, J.~You, and A.~C. Bovik,
\newblock ``Referenceless prediction of perceptual fog density and perceptual
  image defogging,''
\newblock {\em IEEE Transactions on Image Processing}, vol. 24, no. 11, pp.
  3888--3901, 2015.

\bibitem{ferzli2009no}
R.~Ferzli and L.~J. Karam,
\newblock ``A no-reference objective image sharpness metric based on the notion
  of just noticeable blur (jnb),''
\newblock {\em IEEE transactions on image processing}, vol. 18, no. 4, pp.
  717--728, 2009.

\bibitem{hautiere2011blind}
N.~Hauti{\`e}re, J.~P. Tarel, D.~Aubert, and E.~Dumont,
\newblock ``Blind contrast enhancement assessment by gradient ratioing at
  visible edges,''
\newblock {\em Image Analysis \& Stereology}, vol. 27, no. 2, pp. 87--95, 2011.

\bibitem{fog}
``Image \& video quality assessment at live,''
  \url{http://live.ece.utexas.edu/research/fog/},
\newblock Accessed: 2017-10-12.

\end{thebibliography}
\end{small}
\end{document}